\documentclass{article}

\usepackage{arxiv}

\usepackage[utf8]{inputenc} 
\usepackage[T1]{fontenc}    
\usepackage{hyperref}       
\usepackage{url}            
\usepackage{booktabs}       
\usepackage{amsfonts}       
\usepackage{nicefrac}       
\usepackage{microtype}      
\usepackage{lipsum}
\usepackage{graphicx}
\usepackage{amsmath,amssymb} 
\usepackage{color}
\usepackage{algorithm2e}
\usepackage{float}

\title{Iterative Greedy Matching for 3D Human Pose Tracking from Multiple Views}

\author{
 Julian Tanke \\
  University of Bonn \\
  \texttt{tanke@iai.uni-bonn.de} \\
   \And
 J\"urgen Gall \\
  University of Bonn \\
  \texttt{gall@iai.uni-bonn.de} \\
}

\begin{document}
\maketitle
\begin{abstract}
	% ... 

In this work we propose an approach for estimating 3D human poses of multiple people
from a set of calibrated cameras. Estimating 3D human poses from 
multiple views has several compelling properties: human poses are estimated within a 
global coordinate space and 
multiple cameras provide an extended field of view which helps in resolving
ambiguities, occlusions and motion blur.
Our approach builds upon a real-time 2D multi-person pose estimation system and
greedily solves the association problem between multiple views.
We utilize
bipartite matching to track multiple people over multiple frames.
This proofs to be especially efficient as problems associated with greedy matching
such as occlusion can be easily resolved in 3D.
Our approach achieves state-of-the-art results on popular benchmarks and may
serve as a baseline for future work.
\end{abstract}


\section{Introduction}

3D human pose tracking has applications in
surveillance~\cite{zheng2015scalable} and analysis 
of sport events~\cite{burenius20133d,kazemi2013multi}.
Most existing
approaches~\cite{iqbal2018dual,iqbal2018hand,DBLP:conf/bmvc/KostrikovG14,liu2011markerless,martinez2017simple,pavlakos2017coarse,tome2017lifting,mehta2017monocular,mehta2018single}
address 3D human pose estimation from single images
while multi-view 3D human pose 
estimation~\cite{burenius20133d,kazemi2013multi,belagiannis20143d,belagiannis20163d,elhayek2015efficient} 
remains less explored, as obtaining and maintaining 
a configuration of calibrated cameras is difficult and costly.
However, in sports or surveillance, 
calibrated multi-camera setups are available and can be leveraged 
for accurate human pose estimation and tracking.
Utilizing multiple views has several obvious advantages over monocular 3D human
pose estimation: ambiguities introduced by foreshortening as well as
body joint occlusions or motion blurs can be resolved
using other views. Furthermore, human poses are estimated within a 
global coordinate system when using calibrated
cameras. 


\begin{figure}[t]
    \centering
    \includegraphics[width=1\linewidth]{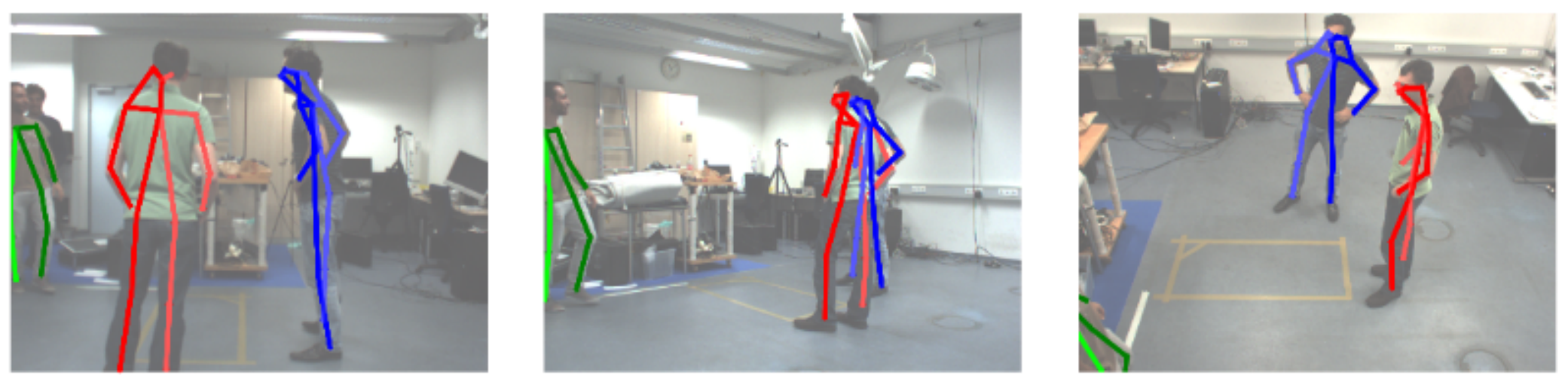}
    \caption{
    Qualitative results on the Shelf~\cite{belagiannis20143d} dataset.
    }
    \label{fig:my_label}
\end{figure}

In this work we propose an iterative greedy matching algorithm 
based on epipolar geometry
to approximately 
solve the k-partite matching problem of multiple human detections in multiple cameras. 
To this end we utilize a real-time 2D pose estimation framework
and achieve very strong
results on challenging multi-camera datasets.
The common 3D space proves to be very robust for greedy tracking, resulting in a very efficient and well-performing
algorithm.
In contrast to previous works~\cite{burenius20133d,kazemi2013multi,pavlakos17harvesting,ershadi2018multiple}, our approach does not discretize the solution space but combines triangulation with an efficient pose association approach across camera views and time.
Furthermore, our approach does not utilize individual shape models for each
person~\cite{liu2011markerless}.



We make the following contributions: 
(i) we present a greedy approach for 3D multi-person tracking
from multiple calibrated cameras and show that our approach achieves state-of-the-art results.
(ii) We provide extensive experiments on both 3D human pose estimation and
on 3D human pose tracking on various 
multi-person multi-camera datasets.

\section{Related Work}

Significant progress has been made in pose estimation and pose tracking in recent 
years~\cite{cao2017realtime,doering2018joint,Iqbal_CVPR2017,xiao2018simple}
and our model is built on advancements in the field of 
2D multi-person pose estimation~\cite{cao2017realtime,chen2018cascaded,fieraru2018learning,guo2018multi,kocabas2018multiposenet,newell2017associative,rogez2019lcr,xiao2018simple}.
For instance, part affinity 
fields~\cite{cao2017realtime} are 2D vector fields that represent associations between body joints which form limbs.
It utilizes a greedy bottom-up approach to detect 2D human poses and is robust to early commitment. 
Furthermore, it decouples the runtime complexity from the number of people in the image, yielding real-time performance.

There is extensive research in monocular 3D human pose 
estimation~\cite{iqbal2018dual,iqbal2018hand,DBLP:conf/bmvc/KostrikovG14,martinez2017simple,pavlakos2017coarse,tome2017lifting,mehta2017monocular,mehta2018single}. For instance,
Martinez et al.~\cite{martinez2017simple} split the problem of 
inferring 3D human poses from single images into 
estimating a 2D human pose and then regressing the 3D pose on
the low-dimensional 2D representation. Though 3D human pose 
estimation approaches from single images yield impressive results
they do not generalize well to unconstrained data.

\begin{figure}[t]
  \begin{center}
  \includegraphics[width=0.7\linewidth]{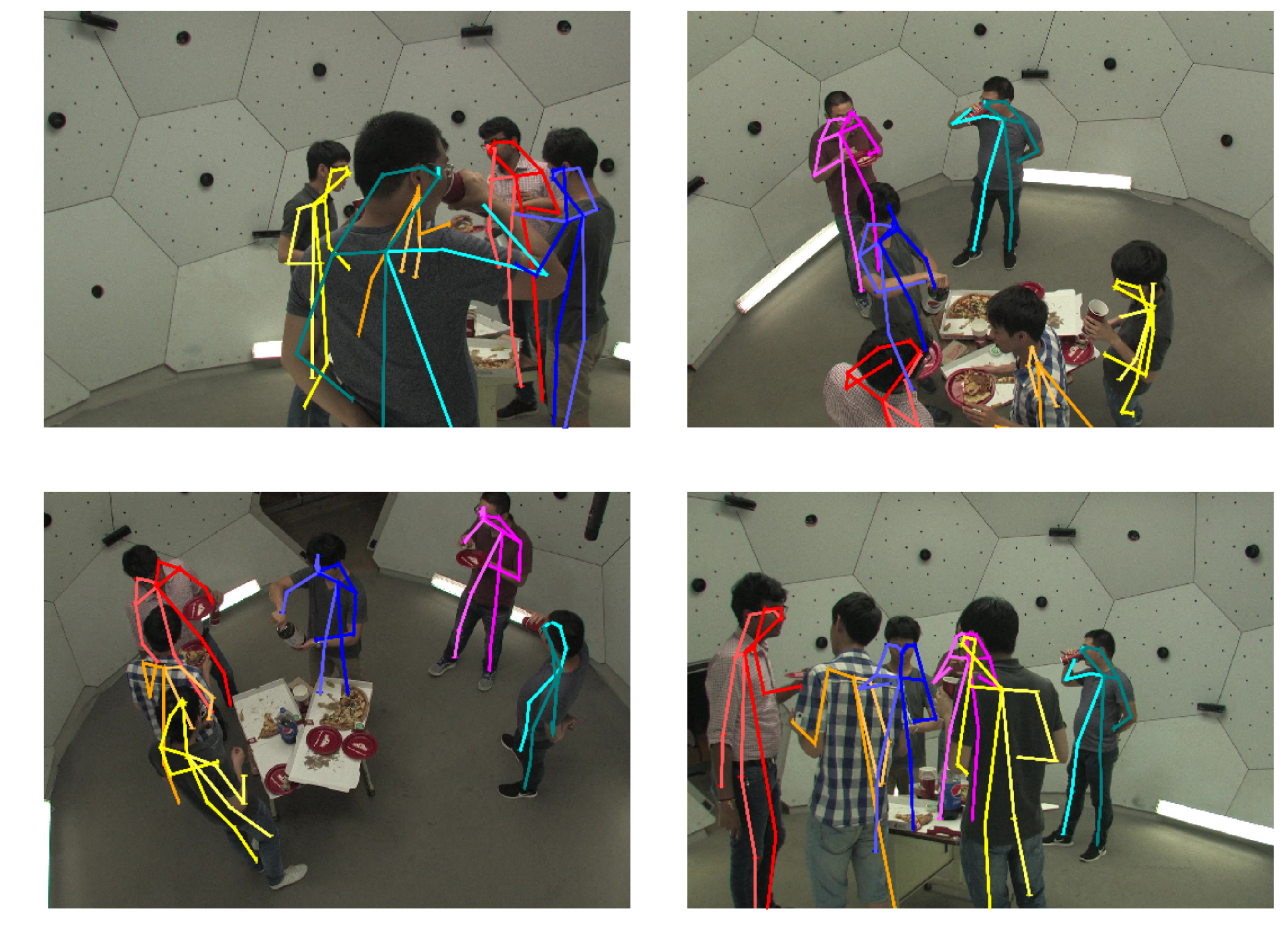}
\end{center}
\caption{Challenging 3D reconstruction of 6 persons in the \textit{CMU Panoptic Dataset}~\cite{Joo_2015_ICCV} with significant occlusion and partial visibility of persons.}
\label{fig:cmu_pizza}
\end{figure}

While multiple views are used in~\cite{pavlakos17harvesting,rhodin2018learning} to guide
the training for monocular 3D pose estimation, there are also approaches that
use multiple views for inference.
A common technique to estimate a single 3D human pose 
from multiple views is to extend the well-known pictorial 
structure model~\cite{felzenszwalb2005pictorial}
to 3D~\cite{amin2013multi,bergtholdt2010study,burenius20133d,kazemi2013multi,pavlakos17harvesting}. Burenius et al.~\cite{burenius20133d} utilize
a 2D part detector based on the HOG-descriptor~\cite{dalal2005histograms}
while Kazemi et al.~\cite{kazemi2013multi} use random forests.
Pavlakos et al.~\cite{pavlakos17harvesting} outperform all
previous models by utilizing the stacked hourglass network~\cite{newell2016stacked} to extract human joint
confidence maps from the camera views.
However, these models have to discretize their solution space resulting
in either a very coarse result or a very large state space making them
impractical for estimating 3D poses of multiple people.
Furthermore, they restrict their solution space to a 3D bounding
volume around the subject which has to be known in advance.
Estimating multiple humans from multiple views was first explored
by Belagiannis et al.~\cite{belagiannis20143d,belagiannis20163d}.
Instead of sampling from all possible translations and rotations
they utilize a set of 3D body joint hypotheses which were obtained
by triangulating 2D body part detections from different views.
However, these methods rely on localizing bounding boxes using 
a person tracker for each individual in each frame
to estimate the number of persons that has to be inferred from
the common state space. This will work well in cases where
individuals are completely visible in most frames but will run
into issues when the pose is not completely visible in some 
cameras as shown in Figure \ref{fig:cmu_pizza}.
A CNN-based approach was proposed by
Elhayek et al.~\cite{elhayek2015efficient} where they fit articulated
skeletons using 3D sums of Gaussians~\cite{stoll2011fast}
and where body part detections are estimated using CNNs.
However, the Gaussians and skeletons 
need to be initialized beforehand
for each actor in the scene, similar to~\cite{liu2011markerless}.
Fully connected pairwise conditional random fields~\cite{ershadi2018multiple}
utilize approximate inference to extract multiple human poses where
DeeperCut~\cite{insafutdinov2016deepercut} is used as 2D human pose estimation model.
However, the search space has to be discretized and a fully connected graph
has to be solved, which throttles inference speed.
Our approach does not suffer from any of the aforementioned drawbacks
as 
our model works off-the-shelf without the need of
actor-specific body models
or discretized state space
and uses an efficient
greedy approach for estimating 3D human poses.

\section{Model}

\begin{figure}[t]
    \centering
    \includegraphics[width=0.6\linewidth]{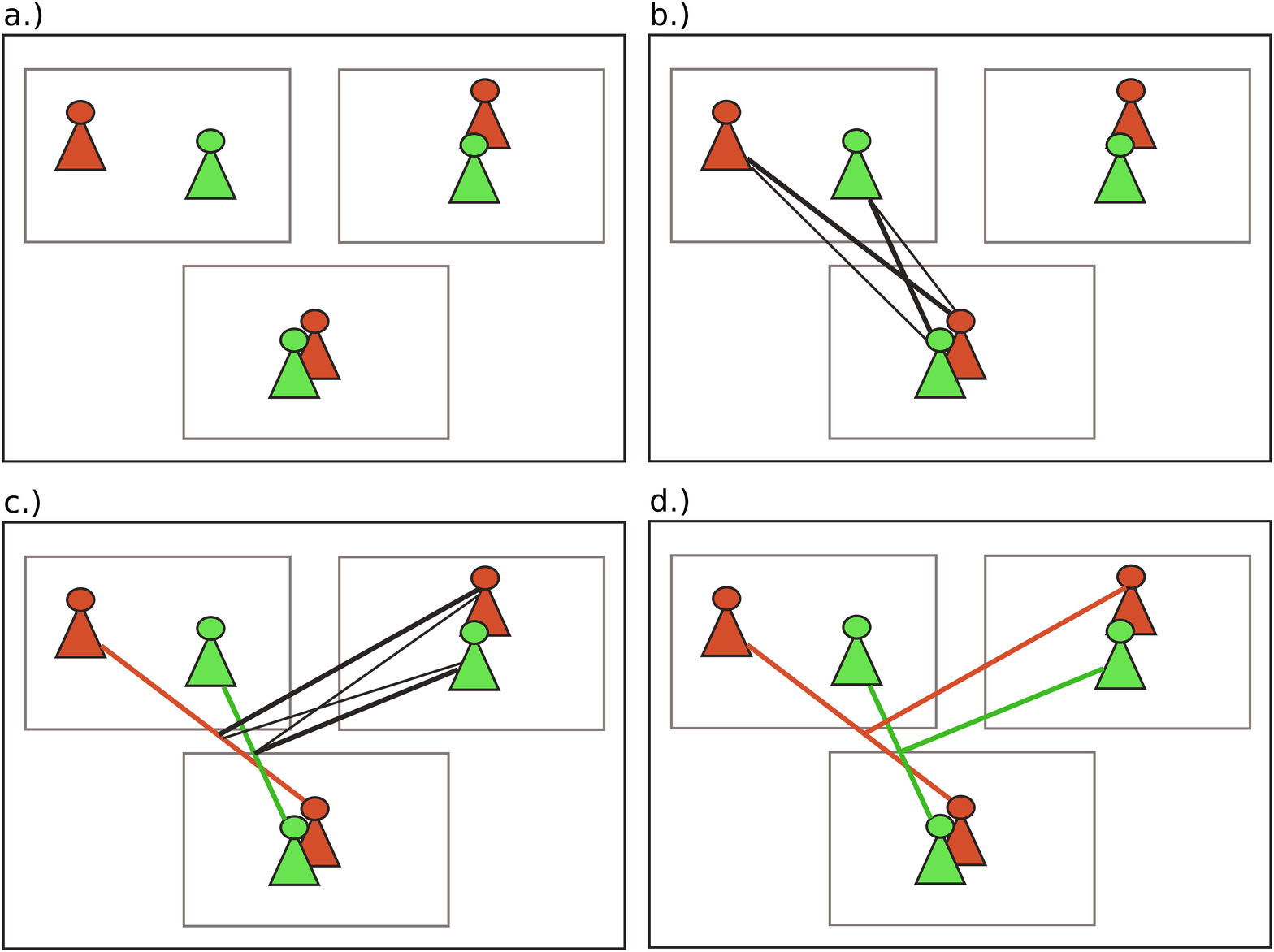}
    \caption{Estimating multiple people from multiple views can be formulated as k-partite graph partitioning
    where 2D human pose detections must be associated across multiple views.
    We employ a greedy approach to make the partitioning tractable. Given a set of 2D human pose detections
    on multiple views (a) we greedily match all detections on two images (b) where the weight between
    two detections is defined by the average epipolar distance of the two poses.
    Other views are then integrated iteratively where the weight is the average of the epipolar distance
    of the 2D detections in the new view and the already integrated 2D detections (c).
    2D detections with the same color represent the same person.
    }
    \label{fig:kpartite_greedy}
\end{figure}


Our model consists of two parts:
First, 3D human poses are estimated for each frame.
Second, the estimated 3D human poses 
are greedily matched into tracks which is described in Section \ref{sec:tracking}. To remove outliers
and to fill-in missing joints in some frames, a simple yet effective 
smoothing scheme is applied, which is also discussed in Section \ref{sec:tracking}.

\subsection{3D Human Pose Estimation}
\label{sec:pose3d}

\begin{figure}[t]
  \begin{center}
  \includegraphics[width=1\linewidth]{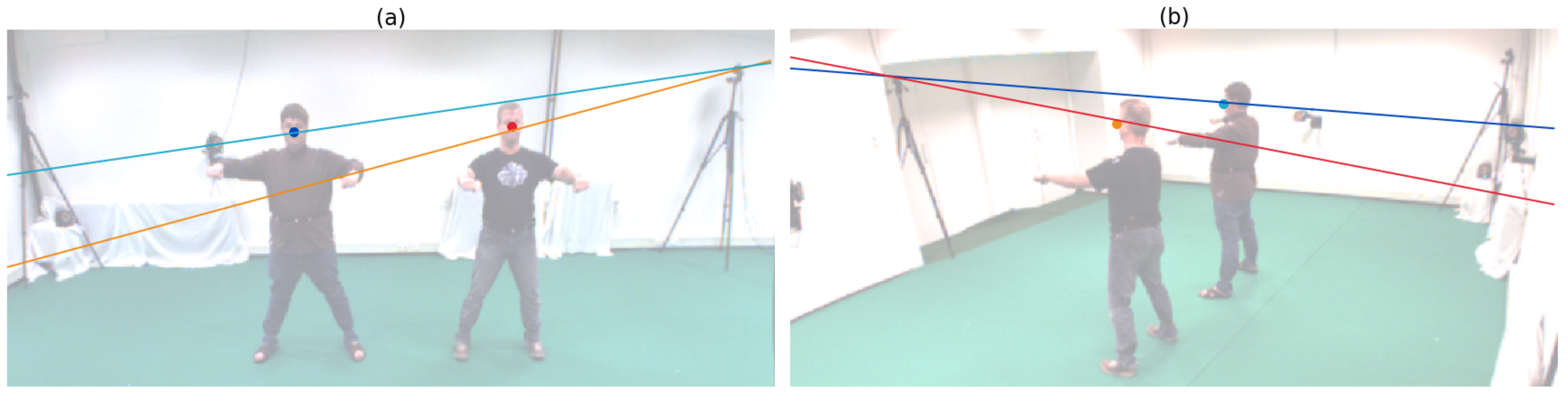}
\end{center}
\caption{Epipolar lines for two camera views of the UMPM Benchmark~\cite{HICV11:UMPM}. 
  The blue and the red dot in image (a) are projected as blue (red) epipolar lines
  in the second image (b) while the orange and light-blue dot from image (b) are 
  projected onto image (a).
  }
\label{fig:epilines}
\end{figure}

First, 2D human poses are extracted for each camera separately.
Several strong 2D multi-person 
pose estimation~\cite{cao2017realtime,chen2018cascaded,fieraru2018learning,guo2018multi,kocabas2018multiposenet,newell2017associative,rogez2019lcr,xiao2018simple} 
models have been proposed but in our baseline we utilize OpenPose~\cite{cao2017realtime} as it is 
well established and offers real-time capabilities.
We denote the 2D human pose estimations as
\begin{align}
\big\{h_{i, k}\big\}_{i \in [1, N]}^{k \in [1, K_i]}    
\end{align}
where $N$ is the number of calibrated cameras and $K_i$ the number of
detected human poses for camera $i$.

In order to estimate the 3D human poses from multiple cameras, we first associate the detections across all views as illustrated in Figure~\ref{fig:kpartite_greedy}. We denote the associated 2D human poses as $\mathcal{H}$ where $\vert \mathcal{H} \vert$ is the number of detected persons and $\mathcal{H}_m = \{h_{i,k}\}$ is the set of 2D human poses that are associated to person $m$. Once the poses are associated, we estimate the 3D human poses for all detected persons $m$ with $\vert \mathcal{H} \vert>1$ by triangulating the 2D joint positions.   

For the association, we select camera $i=1$ as starting point and choose all 2D human
pose detections $h_{1,k}$ in this camera view as person candidates, i.e., 
$\mathcal{H}= \big\{ \{h_{1,k}\} \big\}$. We then iterate over the other cameras and greedily match their 2D detections with the current list of person candidates $\mathcal{H}$ 
using bi-partite matching~\cite{munkres1957algorithms}. 

The cost for assigning a pose $h_{i,k}$ to an existing person candidate $\mathcal{H}_m$ is given by   
\begin{align}
    \Phi(h_{i,k}, \mathcal{H}_m) = \frac{1}{\vert \mathcal{H}_m \vert \vert J_{kl}\vert}
    \sum_{h_{j,l} \in \mathcal{H}_m}
    \sum_{\iota\in J_{kl}}
    \phi(h_{i,k}(\iota), h_{j,l}(\iota))
    \label{eq:epidistance_big}
\end{align}
where $h_{i,k}(\iota)$ denotes the 2D pixel location of joint $\iota$ of the 
2D human pose $h_{i,k}$ and $J_{kl}$ is the set of joints that are visible for both poses $h_{i,k}$ and $h_{j,l}$. Note that the 2D human pose detections might not 
contain all $J$ joints due to occlusions or truncations. The distance between two joints in the respective cameras is defined by the distance between the epipolar lines and the joint locations: 
\begin{align}
    \phi(p_i, p_j) =
\vert p_j^T F^{i,j}p_i \vert + 
    \vert p_i^T F^{j,i}p_j \vert 
    \label{eq:distance}
\end{align}
where $F^{i,j}$ is the fundamental matrix from camera $i$ to 
camera $j$. Figure \ref{fig:epilines} shows the epipolar lines for two joints.

Using the cost function $\Phi(h_{i,k}, \mathcal{H}_m)$, we solve the bi-partite matching problem for each image $i$:
\begin{equation}
X^* = \underset{X}{
 \mathrm{argmin}}\sum_{m=1}^{\vert \mathcal{H} \vert} \sum_{k=1}^{K_i} \Phi(h_{i,k}, \mathcal{H}_m) X_{k,m} 
\end{equation}
where 
\begin{equation}
\nonumber \sum_k X_{k,m} = 1 \; \forall m \quad\text{and}\quad \sum_m X_{k,m} =1 \; \forall k. 
\end{equation}
$X_{k,m}^*=1$ if $h_{i,k}$ is associated to an existing person candidate $\mathcal{H}_m$ and it is zero otherwise. 
If $X_{k,m}^*=1$ and $\Phi(h_{i,k}, \mathcal{H}_m) < \theta$, the 2D detection $h_{i,k}$ is added to $\mathcal{H}_m$. If $\Phi(h_{i,k}, \mathcal{H}_m) \geq \theta$, $\{h_{i,k}\}$ is added as hypothesis for a new person to $\mathcal{H}$.
Algorithm \ref{alg:poseestimation} summarizes the greedy approach for associating the human poses across views.

\begin{algorithm}[t]
\footnotesize
\SetAlgoLined
\KwResult{Associated 2D poses $\mathcal{H}$}
 $\mathcal{H}:= \big\{ \{h_{1,k}\} \big\}$ \;
 \For{\textrm{camera} $i \gets 2$ \KwTo $N$}{
 \For{\textrm{pose} $k \gets 1$ \KwTo $K_i$ }{
    \For{\textrm{hypothesis} $m \gets 1$ \KwTo $\vert\mathcal{H}\vert$}{
    $C_{k,m} = \Phi(h_{i,k}, \mathcal{H}_m)$ \;
    }
 }
 $X^* = \underset{X}{
 \mathrm{argmin}}\sum_{m=1}^{\vert \mathcal{H} \vert} \sum_{k=1}^{K_i} C_{k,m} X_{k,m}$ \;
 \For{$k,m$ \textbf{where} $X_{k,m}^*=1$}{
 \eIf{$C_{k,m} < \theta$}{
 $\mathcal{H}_m = \mathcal{H}_m \ \bigcup \ \{h_{i,k}\}$ \;
 }{
 $\mathcal{H} = \mathcal{H} \ \bigcup \ \big\{ \{h_{i,k} \} \big\}$ \;
 }
 }
 }
 $\mathcal{H} = \mathcal{H} \setminus \mathcal{H}_m \ \forall m$ \textbf{where} $\vert\mathcal{H}_m\vert = 1$\;
 \caption{Solving the assignment problem for multiple 2D human pose detections
 in multiple cameras. $\Phi(h_{i,k}, \mathcal{H}_m)$ \eqref{eq:epidistance_big} is the assignment cost for assigning the 2D human pose $h_{i,k}$ to the person candidate $\mathcal{H}_m$. $X^*$ is a binary matrix obtained by solving the bi-partite matching problem. 
 The last line in the algorithm ensures
 that all hypotheses that cannot be triangulated are removed.
 }
 \label{alg:poseestimation}
\end{algorithm}

\subsection{Tracking}
\label{sec:tracking}

For tracking, we use bipartite matching~\cite{munkres1957algorithms} similar to Section~\ref{sec:pose3d}. Assuming that we have already tracked the 3D human poses until frame $t-1$, we first estimate the 3D human poses for frame $t$ as described in 
Section~\ref{sec:pose3d}. The 3D human poses of frame $t$ are then associated to the 3D human poses of frame $t-1$ by bipartite matching. The assignment cost for two 3D human poses is in this case given by the average Euclidean distance between all joints that are present in both poses. In some cases, two poses do not have any overlapping valid joints due to noisy detections or truncations. The assignment cost is then calculated by projecting the mean of all valid joints of each pose
onto the $xy$-plane, assuming that the $z$-axis is the normal of the ground plane,  
and taking the
Euclidean distance between the projected points. As long as the distance
between two matched poses is below a threshold $\tau$, they will be integrated
into a common track. Otherwise, a new track is created.  
In our experiments we set $\tau=200mm$.


Due to noisy detections, occlusions or motion blur, some joints or even
full poses might be missing in some frames or noisy.
We fill in missing joints by temporal averaging and
we smooth each joint trajectory by a Gaussian kernel with standard deviation $\sigma$. This simple approach significantly boosts the performance of our model as we will show in Section \ref{sec:exp}.

\section{Experiments}

\begin{table}[t]
\normalsize
\begin{center}
    \begin{tabular}{c|| c c c | c c c c c}
    & \cite{burenius20133d}\textsuperscript{*} & \cite{kazemi2013multi}\textsuperscript{*} & \cite{pavlakos17harvesting}\textsuperscript{*} & \cite{belagiannis20143d} & \cite{belagiannis20163d} & 
    \cite{ershadi2018multiple} &
    Ours & Ours\textsuperscript{+} \\
    \hline 
        ua & .60 & .89 & 1.0 & .68 & .98  & .97 & .99 & 1.0 \\
        la & .35 & .68 & 1.0 & .56 & .72  & .95 & .99 & 1.0 \\
        ul & 1.0 & 1.0 & 1.0 & .78 & .99  & 1.0 & .98 & .99 \\
        ll & .90 & .99 & 1.0 & .70 & .92  & .98 & .93 & .997 \\
        \hline
        avg & .71 & .89 & 1.0 & .68 & .90 & .98 & .97 & \textbf{.997} \\
        \hline
    \end{tabular}
\end{center}
\caption{Quantitative comparison of methods for
single human 3D pose estimation from multiple views on
the KTH Football II~\cite{kazemi2013multi} dataset. 
The numbers are the PCP score in 3D with $\alpha=0.5$.
Methods
annotated with \textsuperscript{*} can only estimate single
human poses, discretize the state space and rely on being provided with a tight 3D bounding
box centered at the true 3D location of the person. 
\textit{Ours}\textsuperscript{+} and \textit{Ours}
describe our method with and without 
track smoothing (Section \ref{sec:tracking}). \textit{ul} and
\textit{la} show the scores for upper and lower arm, respectively,
while \textit{ul} and \textit{ll} represent upper and lower legs.
}
\label{tab:singlehuman}
\end{table}

\begin{table}[t]
\normalsize
    \begin{center}
    \begin{tabular}{c|c c c | c c c | c c c | c c c | c c c }
    \multicolumn{13}{c}{\textbf{Campus dataset} $\ (\alpha = 0.5)$} \\
        & \multicolumn{3}{c}{\cite{belagiannis20143d}} 
      & \multicolumn{3}{c}{\cite{belagiannis20163d}} 
      & \multicolumn{3}{c}{\cite{ershadi2018multiple}} 
      & \multicolumn{3}{c}{Ours}  
      & \multicolumn{3}{c}{Ours\textsuperscript{+}}\\ 
      \hline
        Actor & 1 & 2 & 3 & 1 & 2 & 3 & 1 & 2 & 3 & 1 & 2 & 3 & 1 & 2 & 3 \\
        \hline
        ua & .83 &  .90 & .78 & .97 & .97 & .90 
        & .97 & .94 & 93
        & .86 & .97 & .91
        & .99 & .98 & .98\\
   la & .78 &  .40 & .62 & .86 & .43 & .75 
        & .87 & .79 & 70
        & .74 & .64 & .68  
        & .91 & .70 & .92 \\
   ul & .86 & .74 & .83 & .93 & .75 & .92 
        & .94 & .99 & 88 
        & 1.0 & .99 & .99 
        & 1.0 & .98 & 1.0\\
   ll & .91 & .89 & .70 & .97 & .89 & .76 
        & .97 & .95 & 81
        &  1.0 & .98 & .99 
        &  1.0 & .98 & .99 \\
   \hline
   avg & .85 & .73 & .73 & .93 & .76 & .83 
        & .94 & .93 & .85 
        & .90 & .90 & .89 
        & .98 & .91 & .98 \\
   \hline
   avg\textsuperscript{*} & \multicolumn{3}{c|}{.77} & \multicolumn{3}{c|}{.84} 
   & \multicolumn{3}{c|}{.91} 
   & \multicolumn{3}{c|}{.90} 
   & \multicolumn{3}{c}{\textbf{.96}}\\
   \hline
   \multicolumn{13}{c}{} \\
   \multicolumn{13}{c}{\textbf{Shelf dataset}  $\ (\alpha = 0.5)$} \\
        & \multicolumn{3}{c}{\cite{belagiannis20143d}} 
      & \multicolumn{3}{c}{\cite{belagiannis20163d}} 
      & \multicolumn{3}{c}{\cite{ershadi2018multiple}} 
      & \multicolumn{3}{c}{Ours}  
      & \multicolumn{3}{c}{Ours\textsuperscript{+}}\\ 
      \hline
        Actor & 1 & 2 & 3 & 1 & 2 & 3 & 1 & 2 & 3 & 1 & 2 & 3 & 1 & 2 & 3 \\
        \hline
        ua & .72 &  .80 & .91 & .82 & .83 & .93 
        & .93 & .78 & .94
        & .99 & .93 & .97
        &.1.0 & .97 & .97 \\
   la & .61 &  .44 & .89 & .82 & .83 & .93 
        & .83 & .33 & .90 
        & .97 & .57 & .95
        & .99 & .64 & .96 \\
   ul & .37 & .46 & .46 & .43 & .50 & .57 
        & .96 & .95 & .97
        & .998 & 1.0 & 1.0 
        & 1.0 & 1.0 & 1.0 \\
   ll & .71 & .72 & .95 & .86 & .79 & .97 
        & .97 & .93 & .96
        & .998 & .99 & 1.0
        & 1.0 & 1.0 & 1.0 \\
   \hline
   avg & .60 & .61 & .80 & .73 & .74 & .85 
        & .92 & .75 & .94 
        & .99 & .87 & .98
        & .998 & .90 &  .98 \\
   \hline
   avg\textsuperscript{*} & \multicolumn{3}{c|}{.67} & \multicolumn{3}{c|}{.77} 
   & \multicolumn{3}{c|}{.87} 
   & \multicolumn{3}{c|}{.95} 
   & \multicolumn{3}{c}{\textbf{.96}}\\
   \hline
    \end{tabular}
    \end{center}
    \caption{Quantitative comparison of multi-person 3D pose estimation
    from multiple views on the evaluation frames of the
    annotated Campus~\cite{fleuret2007multicamera,belagiannis20143d} and Shelf dataset~\cite{belagiannis20143d}. The numbers are the
    PCP score in 3D with $\alpha=0.5$. \textit{Ours}\textsuperscript{+} and \textit{Ours}
describe our method with and without 
track smoothing (Section \ref{sec:tracking}). We show results
for each of the three actors separately as well as averaged for 
each method (\textit{average}\textsuperscript{*}).
    }
    \label{tab:multihumans}
\end{table}

\begin{table}[t]
\normalsize
    \begin{center}
    \begin{tabular}{c| c c }
    & \multicolumn{2}{c}{Ours\textsuperscript{+}} \\
    \hline
    Actor & 1 & 2 \\
    \hline
    ua & .997 & .98 \\
    la & .98 & .996 \\
    ul & 1.0 & 1.0 \\
    ll & .99 & .997 \\
    \hline
    avg & 0.99 & 0.99 \\
    \hline
    \end{tabular}
    \end{center}
    \caption{
    Quantitative comparison of multi-person 3D pose estimation
    from multiple views on \textit{p2\_chair\_2} of the UMPM benchmark~\cite{HICV11:UMPM}.
    }
    \label{tab:umpm}
\end{table}

\label{sec:exp}

We evaluate our approach on two human pose estimation tasks, single person 3D pose 
estimation and multi-person 3D pose estimation, and compare it to state-of-the-art
methods.
Percentage of correct parts (PCP) in 3D
as described in \cite{burenius20133d} is used for evaluation.
We evaluate on the limbs only as annotated head poses vary significantly throughout various datasets.
In all experiments, the order in which the cameras are processed is given by the dataset.
We then evaluate the tracking performance.
The source code is made publicly available~\footnote{https://github.com/jutanke/mv3dpose}.

\subsection{Single Person 3D Pose Estimation}

Naturally, first works on 3D human pose estimation from multiple
views cover only single humans.
Typical methods~\cite{burenius20133d,kazemi2013multi,pavlakos17harvesting}
find a solution over the complete discretized
state space which is intractable for multiple persons.
However, we report their results for completeness.
All models were evaluated on the complete
first sequence of the second player of the 
KTH Football II~\cite{kazemi2013multi} dataset. Our results are
reported in Table \ref{tab:singlehuman}.
Our model outperforms all other multi-person approaches 
and gets close to the state-of-the-art for 
single human pose estimation~\cite{pavlakos17harvesting} 
which makes strong assumptions
and is much more constrained.
Our model has the lowest accuracy for lower legs (\textit{ll}) which experience 
strong deformation and high movement speed. This can be mostly attributed to
the 2D pose estimation framework which confuses left and right under motion blur,
as can be seen in Figure \ref{fig:earlycommit}.
When smoothing the trajectory (Section \ref{sec:tracking}) this kind of
errors can be reduced.

\subsection{Multi-Person 3D Pose Estimation}

To evaluate our model on multi-person 3D pose estimation, we 
utilize the Campus~\cite{fleuret2007multicamera,belagiannis20143d}, 
Shelf~\cite{belagiannis20143d}, CMU Panoptic~\cite{Joo_2015_ICCV} and UMPM~\cite{HICV11:UMPM} dataset.
The difficulty of the Campus dataset lies in its low resolution 
($360\times 288$ pixel) which makes accurate joint detection hard.
Furthermore, small errors in triangulation or detection will result
in large PCP errors as the final score is calculated on the 3D joint
locations. As in previous works~\cite{belagiannis20143d,belagiannis20163d}
we utilize frames $350-470$ and frames $650-750$ of the Campus dataset
and frames $300-600$ for the Shelf dataset.
Clutter and humans occluding each others make the Shelf dataset
challenging. Nevertheless, our model achieves state-of-the-art
results on both datasets by a large margin 
which can be seen in Table \ref{tab:multihumans}.
Table \ref{tab:umpm} reports quantitative results on video \textit{p2\_chair\_2} of the UMPM~\cite{HICV11:UMPM} benchmark.
A sample frame from this benchmark can be seen in Figure \ref{fig:epilines}.
As the background is homogeneous and the human actors maintain a considerable distance to each other the results of our method are quite strong.

\subsection{Tracking}

\begin{table}[t]
\normalsize
    \begin{center}
    \begin{tabular}{ c| c | c }
        \hline
         & Ours &  Ours\textsuperscript{+} \\
         \hline
        160422\_ultimatum1~\cite{Joo_2015_ICCV} & .89 & .89 \\
        160224\_haggling1~\cite{Joo_2015_ICCV} & .92 & .92 \\
        160906\_pizza1~\cite{Joo_2015_ICCV} & .92 & .93 \\
        \hline
    \end{tabular}    
    \end{center}
    \caption{Quantitative evaluation of multi-person 3D pose tracking on the CMU Panoptic dataset~\cite{Joo_2015_ICCV} using the
    MOTA~\cite{bernardin2006multiple} score.
    \textit{Ours}\textsuperscript{+} and \textit{Ours}
describe our method with and without 
track smoothing (Section \ref{sec:tracking}).
    }
    \label{tab:mota}
\end{table}

For evaluating the tracking accuracy, we utilize the 
MOTA~\cite{bernardin2006multiple} score which provides a scalar value
for the rate of false positives, false negatives, and identity switches
of a track.
Our model is evaluated on the CMU Panoptic dataset~\cite{Joo_2015_ICCV} 
which provides multiple interacting people in close proximity.
We use videos \textit{160224\_haggling1} with three persons,
\textit{160422\_ultimatum1} with up to seven person,
and \textit{160906\_pizza1} with six persons.
For the videos
\textit{160422\_ultimatum1} we use frames $300$ to $3758$, for 
\textit{160906\_pizza1} we use frames $1000$ to $4458$ and 
for \textit{160224\_haggling1} we use frames $4209$ to $5315$ and 
$6440$ to $8200$.
The first five HD cameras are used.
Our results are reported in Table \ref{tab:mota} which shows that
our approach yields strong tracking capabilities.


\subsection{Effects of Smoothing}

\begin{figure}[t]
  \begin{center}
  \includegraphics[width=1\linewidth]{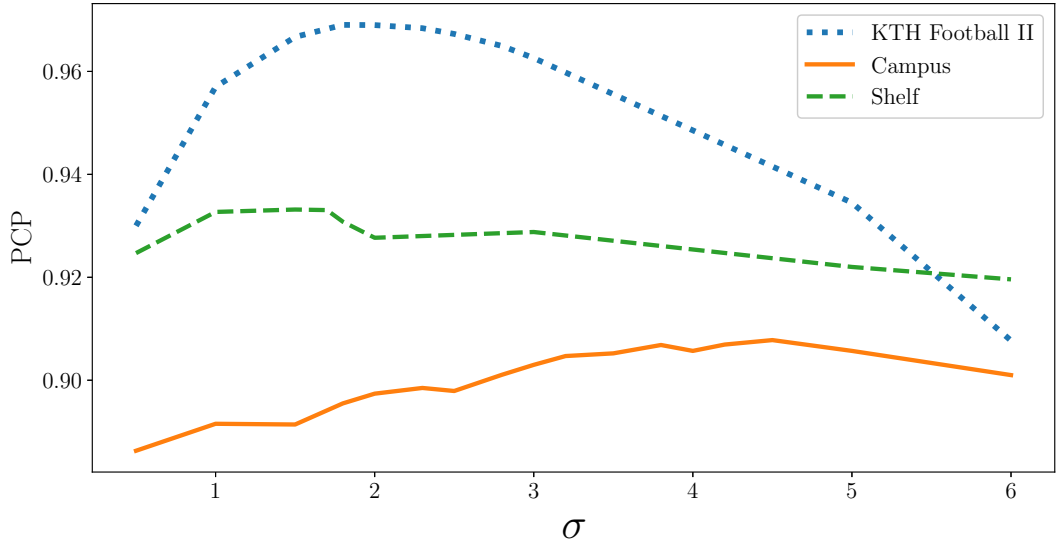}
\end{center}
\caption{
PCP score for different smoothing values $\sigma$ for tracking on KTH Football II,
Campus, and Shelf.
If $\sigma$ is too small, the smoothing has little effect and 
coincides with the un-smoothed results. When the joint trajectories
are smoothed too much, the PCP score drops as well as the trajectories do not follow
the original path anymore. (Larger PCP scores are better)
}
\label{fig:sigma}
\end{figure}

As can be seen in Table \ref{tab:singlehuman} and Table \ref{tab:multihumans}
the effects of smoothing can be significant, especially when detection
and calibration are noisy as is the case with the Campus and the KTH 
Football II dataset. In both datasets 2D human pose detection 
is challenging due to
low resolution (Campus) or strong motion blur (KTH Football II).
Datasets with higher resolution and less motion blur like the Shelf dataset
do not suffer from this problems as much and as such do not benefit
the same way from track smoothing. However, a small gain can still be noted
as smoothing also fills in joint detections that could not be triangulated.
Figure \ref{fig:sigma} explores different $\sigma$ values for smoothing
on the KTH Football II, Campus, and Shelf dataset. 
It can be seen that
smoothing improves the performance regardless of the dataset but that too much
smoothing obviously reduces the accuracy. We chose $\sigma=2$ for all our experiments except for the
Campus dataset where we set $\sigma=4.2$.
The reason for the higher value of $\sigma$ for the Campus dataset is due to the very low resolution of the images compared to the other datasets, which increases the noise of the estimated 3D joint position by triangulation.    

\subsection{Effects of camera order}

\begin{figure}[t]
  \begin{center}
  \includegraphics[width=1\linewidth]{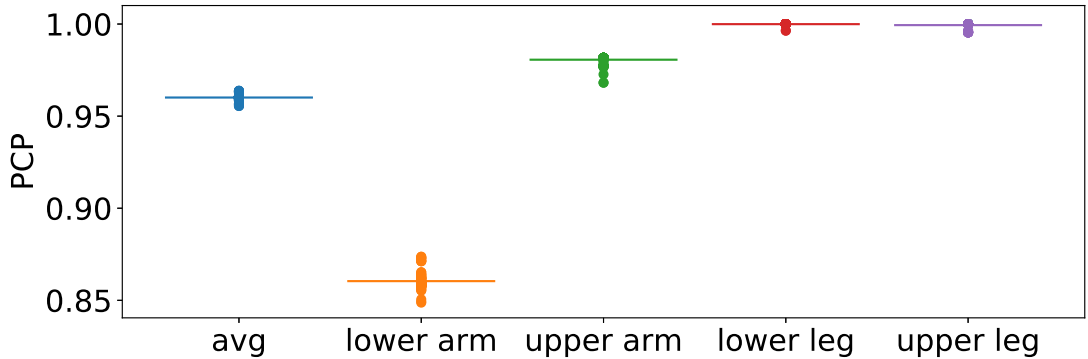}
\end{center}
\caption{
PCP score averaged over all subjects for all $120$ camera permutations of the Shelf dataset.
The vertical line represents the mean value over all permutations while the dots represent each camera permutation.
}
\label{fig:permvar}
\end{figure}

So far we used the given camera order for each dataset, but the order in which views are greedily matched matters and different results might happen with different orderings. To investigate the impact of the camera order, we evaluated our approach using all $120$ permutations of the $5$ cameras of the Shelf dataset. The results shown in Figure \ref{fig:permvar} show that the approach is very robust to the order of the camera views.        



\subsection{Early Commitment}
\begin{figure}[t]
  \begin{center}
  \includegraphics[width=0.75\linewidth]{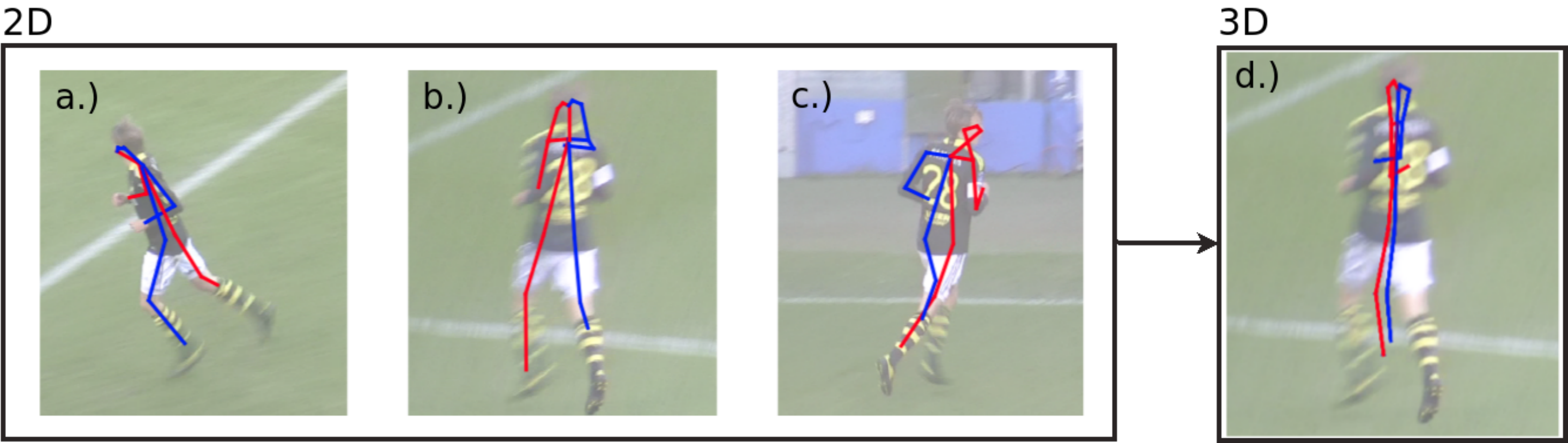}
\end{center}
\caption{
Issues with early commitment. As we utilize the 2D pose estimations
directly, our method suffers when the predictions yield poor results. 
In this example
the pose estimation model correctly estimates (a) and (c) but
confuses left and right on (b) due to motion blur. The resulting 3D pose 
estimation (d) collapses into the centre of the person. The red limbs represent the right
body side while blue limbs represent the left body side.
}
\label{fig:earlycommit}
\end{figure}
A failure case happens due to the early commitment of our 
algorithm with regards to the 2D pose estimation, as can be seen in 
Figure \ref{fig:earlycommit}.
When the pose estimation is unsure about a pose, it still fully commits to
its output and disregards uncertainty. This problem occurs due to motion blur as the network has difficulties to decide between
left and right in this case. As our pose estimation model has mostly seen forward-facing
persons it will be more inclined towards predicting a forward-facing person
in case of uncertainty. When left and right of a 2D prediction 
are incorrectly flipped in at least one of the views, the merged 3D prediction
will collapse to the vertical line of the person resulting in a poor
3D pose estimation.

\section{Conclusion}

In this work we presented a simple baseline approach for 3D human pose estimation 
and tracking
from
multiple calibrated cameras and evaluate it extensively on several 3D multi-camera
datasets. Our approach achieves state-of-the-art results in multi-person 3D pose estimation
while remaining sufficiently efficient for fast processing.
Due to the models simplicity some common failure cases can be noted which
can be build upon in future work. For example, confidence maps provided by
the 2D pose estimation model could be utilized to prevent left-right flips.
Our approach may serve as a baseline for future work.

\section*{Acknowledgement}
The work has been funded by the Deutsche Forschungsgemeinschaft (DFG, 
German Research Foundation) GA 1927/5-1 (FOR 2535 Anticipating Human 
Behavior) and the ERC Starting Grant ARCA (677650).

\bibliographystyle{splncs04}
\bibliography{egbib}

\end{document}